\documentclass[10pt,twocolumn,letterpaper]{article}

\usepackage{iccv}
\usepackage{times}
\usepackage{epsfig}
\usepackage{graphicx}
\usepackage{amsmath}
\usepackage{amssymb}
\usepackage{multirow}
\usepackage{fancyhdr}

\usepackage[breaklinks=true,bookmarks=false]{hyperref}

\iccvfinalcopy 

\def\iccvPaperID{****} 

\begin{document}

\title{Video captioning with recurrent networks based on \\ frame- and
  video-level features and visual content classification}

\author{Rakshith Shetty and Jorma Laaksonen\\
  Department of Computer Science\\
  Aalto University School of Science\\
  P.O.BOX 15400, FI-00076 AALTO\\
  Espoo, Finland \\ 
  {\tt\small rakshith.shetty@aalto.fi \quad jorma.laaksonen@aalto.fi}
}

\maketitle
\thispagestyle{fancy}

\begin{abstract}
  In this paper, we describe the system for generating textual
  descriptions of short video clips using recurrent neural networks,
  which we used while participating in the Large Scale Movie
  Description Challenge 2015 in ICCV 2015. 
  Our work builds on static image captioning system proposed
  in~\cite{Vinyals_2015_CVPR} and implemented
  in~\cite{Karpathy_2015_CVPR} and extends this framework to videos
  utilizing both static image features and video-specific features.
  In addition, we study the usefulness of visual content classifiers
  as a source of additional information for caption generation.
  With experimental results we show that utilizing keyframe based
  features, dense trajectory video features and content classifier
  outputs together gives better performance than any one of them
  individually. We also release the source code for our work online 
  \textsuperscript{\ref{ourCode}}.

\end{abstract}


\section{Introduction}

Automatic description of videos using sentences of natural language is
a challenging task in computer vision.
Recently, two large collections of video clips extracted from movies,
together with natural language descriptions of their visual content,
have been published for scientific purposes.
A unified version of these data sets, namely
M-VAD~\cite{AtorabiM-VAD2015} and MPII-MD~\cite{rohrbach15cvpr}, has
been provided for the purpose of the Large Scale Movie Description
(LSMDC) Challenge 2015.
In this paper, we describe the system we used while participating in
LSMDC 2015%
\footnote{\scriptsize\url{https://sites.google.com/site/describingmovies/}}.

Our work builds on the static image captioning system based on
recurrent neural networks and proposed in~\cite{Vinyals_2015_CVPR} and
implemented in the \emph{NeuralTalk}%
\footnote{\scriptsize\url{https://github.com/karpathy/neuraltalk}}
system~\cite{Karpathy_2015_CVPR}.
We extend this framework for generating textual descriptions of small
video clips utilizing both static image features and video-specific
features.
We also study the use of visual content classifiers as a source of
additional information for caption generation. We also make the source 
code for our work available online%
\footnote{\label{ourCode}\scriptsize\url{https://github.com/aalto-cbir/neuraltalkTheano}}
.


\section{Model}

\subsection{Overview}

\begin{figure}[t]
  \begin{center}
    \includegraphics[width=1.1\linewidth]{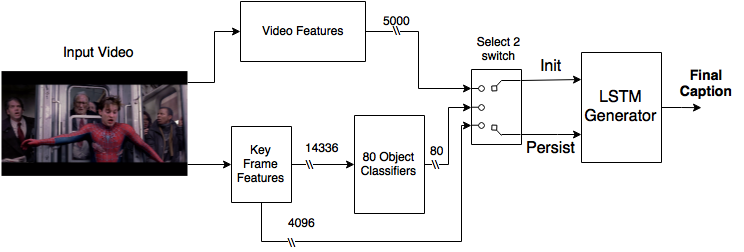}
  \end{center}
  \caption{Block diagram of our model}
  \label{fig:blkdiag}
\end{figure}

We propose to use a neural-network-based framework to generate textual
captions for the given input video. 
Our pipeline consists of three distinct stages as seen in
Figure~\ref{fig:blkdiag}. 
The first stage is feature extraction, wherein we extract both whole
video and keyframe image based features from the input video.
As the whole video based feature we use dense
trajectories~\cite{DBLP:conf/cvpr/WangKSL11,DBLP:journals/ijcv/WangKSL13}. 
Keyframe image features are extracted by feeding these images through
Convolutional Neural Networks (CNN) trained on the ImageNet
database~\cite{imagenet_cvpr09}.
We use three different CNN architectures giving us a rich variety of
features for the keyframe images.
In many studies, including our own~\cite{ACMMM2014,TRECVID2014},
CNN-based features and especially late fusion combinations of them
have been found to provide superior performance in many computer
vision and image analysis tasks.

In our current experiments, the CNN-based features are either directly
input to LSTM, the Long short-term memory~\cite{Hochreiter1997}
recurrent neural network (RNN), or they are fed to a set of visual
content classifiers which in turn produce 80-dimensional class
membership vectors that can then be used as LSTM inputs.
For training the 80 classifiers we have used the training set images
of the COCO 2014 collection~\cite{Lin2014}.

The third stage consists of an LSTM network, taking one feature set
and possibly the visual content classification results as input and
generating a sequence of words, i.e.\@ a video caption, with the
highest probability of being associated with the input features and
thus the processed video clip.
Next we will look at each of these processing stages in detail.


\subsection{Video feature extraction}

As the video-level features we use the dense
trajectories~\cite{DBLP:conf/cvpr/WangKSL11},
\cite{DBLP:journals/ijcv/WangKSL13} which have proven to have good
performance in many video analysis tasks.

We have used the minimum length of 2 seconds for the video clips when
extracting the dense trajectory features.
Videos that had frames wider than 800 pixels we scaled down to the
width of 720 pixels.
We used the default length of 15 frames in the trajectories, giving
rise to 28 dimensional displacement vectors.
These we quantized to a 1000-dimensional histogram, whose codebook was
created with k-means clustering of 1,000,000 randomly sampled
trajectories from the training partition of the LSMDC 2015 video clips.

In addition, we created also 1000-dimensional histograms from the
96-dimensional HOG~\cite{dalal2005}, Motion Boundary MBHx and
MBHy~\cite{Dalal2006} descriptors and the 108-dimensional
HOF~\cite{Laptev2008} descriptors of the dense trajectories.
Concatenating all these five histograms resulted in 5000-dimensional
video features.
In many cases dense trajectory features of higher dimensionality, say
$5\times4096=20480$ have been found to be better than feature vectors
of this dimensionality, but we were afraid that the training of the
LSTM network might not be successful with too high-dimensional inputs.


\subsection{Image feature extraction}

We also extract static image features from one keyframe selected from
the center of each video clip.
For the feature extraction in the keyframes we use CNNs pre-trained on
the ImageNet database for object
classification~\cite{imagenet_cvpr09}.
We use three different CNN architectures namely 16-layer and 19-layer
VGG~\cite{Simonyan14c} nets and
GoogLeNet~\cite{DBLP:journals/corr/SzegedyLJSRAEVR14}.  
In the case of VGG nets we extract the activations of the network on
the second fully-connected 4096-dimensional \emph{fc7} layer for the
given input images whose aspect ratio is distorted to a square.
Ten regions, as suggested in~\cite{Krizhevsky2012}, are extracted
from all images and average pooling of the region-wise features are
used to generate the final features.

For GoogLeNet we have used similarly the \emph{5th Inception module},
having the dimensionality of 1024.
We augment these features with the reverse spatial
pyramid pooling proposed in~\cite{Gong2014} with two scale levels.
The second level consists of a $3\times3$ grid with overlaps and
horizontal flipping, resulting in a total of 26 regions, on the scale
of two.
The activations of the regions are then pooled using average and
maximum pooling.
Finally, the activations of the different scales are concatenated
resulting to 2048-dimensional features.
See \cite{ACMMM2014} for some more details.


\subsection{Visual content classification}

We have extracted the above described five CNN-based image features
also from the images of the COCO 2014~\cite{Lin2014} training set and
trained an SVM classifier for each of the 80 object categories
specified in COCO 2014.
In particular, we utilized linear SVMs with homogeneous kernel
maps~\cite{Vedaldi2010} of order $d=2$ to approximate the intersection
kernel.
Furthermore, we used two rounds of hard negative mining~\cite{Li2013}
and sampled $5\,000$ negative examples on each round.

For each LSMDC keyframes we thus have 15 SVM outputs (five features
times initial and two hard negative trainings) that we combine with
arithmetic mean in the late fusion stage.
The 80 fusion values, one for each object category, are then
concatenated to form a class membership vector for each keyframe
image.
These vectors we optionally use as inputs to the LSTM network.


\subsection{LSTM caption generator}

\begin{figure}[t]
\begin{center}
   \includegraphics[width=0.6\linewidth]{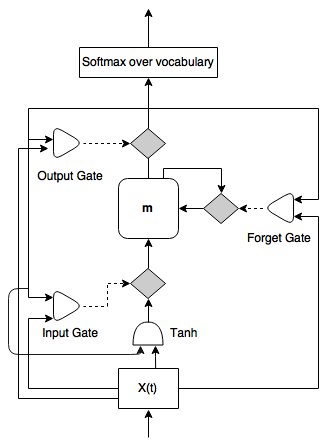}
\end{center}
\caption{Block diagram of a single LSTM cell. dotted lines indicate
  gate controls and full lines are data flow. Triangle indicates
  sigmoid non linearity.}
\label{fig:lstmcell}
\end{figure}

To learn generative models of sentences conditioned on the input
image, video and class membership features, we chose to use LSTM
networks~\cite{Hochreiter1997}.
This choice was based on two basic requirements this problem
imposes. Firstly, the model needs to handle sentences of arbitrary
length and LSTMs are able to do this by design. Secondly, during
training using gradient descent methods the error signal and its
gradients need to propagate a long way back in time without exploding,
and again LSTMs satisfy this criteria.

The block diagram of a single LSTM cell is shown in
Figure~\ref{fig:lstmcell}.  It consists of a memory cell $m$, whose
value at any timestep $t$ is influenced by the current input $x$,
previous output $y$ and previous cell state $m(t-1)$. The update to
the memory value $m$ is controlled using the input gate and the forget
gate.  The output is controlled using the output gate. The gates are
implemented using sigmoidal non-linearity keeping them completely
differentiable. 
The input and forget gates LSTM cells have the ability to preserve the
content of the memory cell over long periods making it easier to learn
longer sequences.
This process is formalized in the equations below:
\begin{align}
i(t) &= \sigma(W_{ix}x(t-1) + W_{iy}y(t-1))\\
o(t) &= \sigma(W_{ox}x(t-1) + W_{oy}y(t-1))\\
f(t) &= \sigma(W_{fx}x(t-1) + W_{fy}y(t-1))\\
\begin{split}
m(t) &= f(t)\cdot m(t-1) + \\
     &\; \; \; \; i(t)\cdot \tanh(W_{mx}x(t)+W_{my}y(t-1))
\end{split}\\
y(t) &= o(t) \cdot m(t) \;.
\end{align}

We add a softmax layer at the output of the LSTM to generate a probability 
distribution over the vocabulary. 
At each time step, the LSTM is trained to assign the highest
probability to the word it thinks should appear next given the current
inputs and the hidden state:
\begin{align}
p(w_t | w_{t-1},\cdots,w_0, V) = \text{softmax}(y(t)) \;.
\end{align}

In our simplest architecture, visual features are fed into the LSTM
only at the zeroth time step as the input $x(0)$.
We refer to this feature input as the \emph{init} feature since it
initializes the hidden state of the LSTM.
In the subsequent time steps, a start symbol followed by the word
embeddings for each word in the reference caption are fed through
input $x(t)$.  
In our experiments with the COCO dataset, we have found it helpful to
let the LSTM have access to the visual features throughout the
generation process. 
This requires adding to the LSTM cell a new input which we refer to as
the \emph{persistent} features.
This input plays the same role as $x(t)$ in equations~(1)--(4) except
with a different set of weights.
Note that we can input different visual features in the \emph{init}
and \emph{persistent} lines thereby allowing the model to learn
simultaneously from two different sources.

The training procedure for the LSTM generator is the same as
in~\cite{Vinyals_2015_CVPR}, we try to maximize the log probability
assigned to the training samples by the model. Equivalently, we can
minimize the negative log likelihood as given as:
\begin{align}
\label{eqCost}
L(w_{1:L} | V) = -\sum_{i=1}^N \log(p(w_i|w_{i-1},V)) \; .
\end{align}


\section{Experiments and results}

\begin{table*}[t]
  \newcommand{\modpar}[4]{%
    \multirow{2}{*}{\emph{#1}} & \multirow{2}{*}{#2} & \multirow{2}{*}{#3}
    & \multirow{2}{*}{#4}}
\centering
    \begin{tabular}{|l|c|c|c|c|c|c|c|c|c|c|}
        \hline\hline
        \bf model    &\bf init &\bf persist &\bf perplex &\bf beam size &\bf avg.len &\bf Bleu\_4 &\bf CIDEr &\bf METEOR &\bf ROUGE\_L    & \\\hline\hline
        \modpar{1 coco-kf}{kf}{--}{ - }      & 1 & 9.62  &   0.003 &   0.049 &   0.053 &   0.116 &     \\\cline{5-11}
                                       & & & & 5 & 8.74  &   0.003 &   0.045 &   0.044 &   0.105 & (p) \\\hline
        \modpar{2 coco-kf+cls}{kf}{cls}{ - } & 1 & 10.17 &   0.003 &   0.045 &   0.053 &   0.113 &     \\\cline{5-11}
                                       & & & & 5 & 9.04  &   0.003 &   0.045 &   0.044 &   0.103 &     \\\hline
        \modpar{3 coco-cls+kf}{cls}{kf}{ - } & 1 & 9.62  &   0.003 &   0.052 &   0.053 &   0.114 &     \\\cline{5-11}
                                       & & & & 5 & 8.66  &   0.003 &   0.049 &   0.045 &   0.104 &     \\\hline\hline
        \modpar{4 kf}{kf}{--}{56.08}         & 1 & 5.24  &   0.004 &   0.071 &\bf0.058 &   0.140 &     \\\cline{5-11}
                                       & & & & 5 & 3.39  &   0.002 &   0.063 &   0.043 &   0.114 &     \\\hline
        \modpar{5 kf+cls}{kf}{cls}{60.78}    & 1 & 5.36  &   0.004 &   0.073 &\bf0.060 &   0.142 &     \\\cline{5-11}
                                       & & & & 5 & 3.46  &   0.001 &   0.054 &   0.043 &   0.111 &     \\\hline
        \modpar{6 cls+kf}{cls}{kf}{59.07}    & 1 & 5.12  &\bf0.005 &   0.087 &\bf0.059 &\bf0.144 &     \\\cline{5-11}
                                       & & & & 5 & 3.50  &   0.003 &   0.071 &   0.047 &   0.122 &     \\\hline\hline
        \modpar{7 traj}{traj}{--}{54.89}     & 1 & 5.28  &\bf0.005 &   0.087 &   0.057 &\bf0.145 &     \\\cline{5-11}
                                       & & & & 5 & 3.75  &   0.003 &   0.074 &   0.047 &   0.123 &     \\\hline
        \modpar{8 traj+cls}{traj}{cls}{59.75}& 1 & 5.28  &\bf0.005 &   0.081 &   0.057 &   0.141 &     \\\cline{5-11}
                                       & & & & 5 & 3.48  &   0.003 &   0.074 &   0.047 &   0.123 &     \\\hline
        \modpar{9 cls+traj}{cls}{traj}{55.14}& 1 & 5.33  &\bf0.006 &\bf0.092 &\bf0.058 &\bf0.146 & (b) \\\cline{5-11}
                                       & & & & 5 & 3.80  &   0.004 &   0.082 &   0.049 &   0.128 &     \\\hline\hline
    \end{tabular}
    \medskip
    \caption{Results obtained on the public test set of LSMDC2015. 
      ``kf'' stands for using keyframe-based features, ``traj'' for
      dense trajectory-based video features and ``cls'' for visual 
      content classification results as inputs to the \emph{init}
      and \emph{persistent} input lines of the LSTM network.
      Submissions (p) and (b) were the ones visible in the
      CodaLab leaderboard for the public and blind test sets, 
      respectively, by the closing time of the LSMDC 2015 Challenge.}
    \label{tab:results}
\end{table*}

To evaluate various forms of our model we used the the LSMDC 2015
public test set as the benchmark. The evaluation is performed using four
standard metrics used in the LSMDC evaluation server namely: 
METEOR~\cite{denkowski-lavie:2014:W14-33},
BLEU~\cite{Papineni:2002:BMA:1073083.1073135}, 
ROUGE-L~\cite{lin2004rouge} and CIDEr~\cite{Vedantam_2015_CVPR}.
Table~\ref{tab:results} shows these four metrics computed for
different models.
In addition to the metrics, we also show the perplexity of the model
on the public test set and the average lengths of the generated
sentences.
Results are provided always for beam sizes 1 and 5 used in the
caption generation stage.

\newcommand{\model}[1]{\emph{#1}}

In order to get a quick baseline, we used models trained earlier on
the COCO dataset to generate captions on the LSMDC test set with a
simple rule-based translation applied to their output.  This
translation is done in order to better match the LSMDC vocabulary and
is implemented using the simple $w_{\text{in}} \longrightarrow w_{\text{out}}$
rule:
\begin{align}
\label{eqTrans}
w_{\text{out}} = \begin{cases}
\text{SOMEONE},& \text{if } w_{\text{in}} \in \{\text{man}, \text{woman}, 
\\& \text{\mbox{\qquad\qquad person}}, \text{boy}, \text{girl} \}\\
w_{\text{in}},& \text{otherwise.}
\end{cases}
\end{align}

Models 1--3 in Table~\ref{tab:results} are such translated models
trained on the COCO dataset.
Model \model{1~coco-kf} was trained on the COCO dataset using
concatenated GoogLeNet-based features with a total dimensionality of
4096 as the \emph{init} features. 
This approach matches the use of the NeuralTalk model described
in~\cite{Vinyals_2015_CVPR} and~\cite{Karpathy_2015_CVPR}. 
Model \model{2~coco-kf+cls} was trained using GoogLeNet as the
\emph{init} features and the outputs of the 80 SVM classifiers as the
\emph{persistent} feature, while model \model{3~coco-cls+kf} was
trained with the role of these two feature types reversed.
The results of these models have in our earlier experiments shown
increasingly better performance on the COCO dataset itself, but we can
hardly observe such progression in the translated results on the LSMDC
dataset.

Next, we trained three models similar to the above COCO models, but now
with captions available and features and content classification
results calculated from the keyframes of the videos in the LSMDC 2015
dataset.
The results are presented as models 4--6 in
Table~\ref{tab:results}. 
Here we can see the benefit of using \emph{persistent} features as the
model \model{6~cls+kf} performs better than the models trained solely
on keyframe features.

Finally, we trained three models using the dense trajectory video
features and the keyframe-based SVM output features, presented in
Table~\ref{tab:results} as models 7--9.
Again we see that using the higher-dimensional feature, here the
dense trajectory feature, as the \emph{persistent} input to the LSTM
network gives the best performance among the group of models.
The result of model \model{9~cls+traj} can also be regarded as the
best one obtained in our experiments and therefore we have used it
in our final blind test data submission to the LSMDC 2015 Challenge.

As we can see from Table 1, the \emph{persistent} dense trajectory
video features combined with the \emph{init} SVM classifier features
from keyframes outperform all the other models in three out of the
four metrics used.
Comparing this with model \model{6~cls+kf} shows that using video
features as opposed to just keyframe features gives a better
performance.
It also indicates that combining the keyframe and video features is
better than using just the video features.

A rather surprising finding is that using larger beam sizes in
inference lead to poorer performance.
This is slightly counterintuitive, but can be understood when we look
at the lengths of the sentences produced by these two beam sizes. 
For example, model \model{9~cls+traj} produces sentences with the
average length of 5.33 words with beam size 1, while with beam size 5
the average length drops to just 3.79 words. This is because with
higher beam sizes the model always picks the most likely sentence and
penalizes heavily any word it is unsure of.  This results in the model
picking very generic sentences like \emph{``SOMEONE looks at
  SOMEONE''} over more descriptive ones.

The results~(p) and~(b) in Table~\ref{tab:results} match our public
and blind test data submissions, respectively, visible in the CodaLab
leaderboard%
\footnote{\scriptsize\url{https://www.codalab.org/competitions/6121\#results}}.


\section{Conclusions}

In this paper, we described the framework of techniques utilized in 
our participation in the LSMDC 2015 Challenge.
We presented a technique which utilizes (1)~video-based dense
trajectory features, (2)~keyframe-based visual features, and
(3)~object classifier output features, and an LSTM network to generate
video descriptions therefrom.
We discussed a couple of architectural variations and experimentally
determined the best architecture among them for this dataset.
We also experimentally verified the effect of the beam size used in
the inference stage on the performance of the captioning system.
The two conclusions we made are:
(1)~Using the classifier output features to initialize the LSTM
network and video features after the initialization results in the
best performance.
(2)~Beam size one in the sentence generation process is better than
larger beam sizes.


\section{Acknowledgments}

This work has been funded by the Academy of Finland (Finnish Centre of
Excellence in Computational Inference Research COIN, 251170) and
\emph{Data to Intelligence (D2I)} DIGILE SHOK project.  The
calculations were performed using computer resources within the Aalto
University School of Science ``Science-IT'' project.


{\small
\bibliographystyle{ieee}
\bibliography{picsom,others}
}

\end{document}